\documentclass[manuscript]{acmart}
\AtBeginDocument{%
  \providecommand\BibTeX{{%
    \normalfont B\kern-0.5em{\scshape i\kern-0.25em b}\kern-0.8em\TeX}}}

\setcopyright{acmcopyright}
\copyrightyear{2022}
\acmYear{2022}
\acmDOI{XXXXXXX.XXXXXXX}

\acmJournal{JACM}
\acmVolume{37}
\acmNumber{4}
\acmArticle{111}
\acmMonth{8}

\usepackage{graphicx}
\graphicspath{ {./images/} }

\usepackage{natbib}
\usepackage{tabularx}

\usepackage{amsthm}

\usepackage{subfig}

\begin{document}

\title{Rationalization for Explainable NLP: A Survey}

\author{Sai Gurrapu}
\email{saig@vt.edu}
\affiliation{%
  \institution{Department of Computer Science, Virginia Tech}
  \country{USA}
}

\author{Ajay Kulkarni}
\email{lakulkar8@gmu.edu}
\affiliation{%
  \institution{Department of Computational and Data Sciences, George Mason University}
  \country{USA}
}

\author{Lifu Huang}
\email{lifuh@vt.edu}
\affiliation{%
  \institution{Department of Computer Science, Virginia Tech}
  \country{USA}
}

\author{Ismini Lourentzou}
\email{ilourentzou@vt.edu}
\affiliation{%
  \institution{Department of Computer Science, Virginia Tech}
  \country{USA}
}

\author{Laura Freeman}
\email{laura.freeman@vt.edu}
\affiliation{%
  \institution{Department of Statistics, Virginia Tech}
  \country{USA}
}

\author{Feras A. Batarseh}
\email{batarseh@vt.edu}
\orcid{0000-0002-6062-2747}
\affiliation{%
  \institution{Department of Electrical and Computer Engineering, Virginia Tech}
  \country{USA}
}

\renewcommand{\shortauthors}{Gurrapu, et al.}

\begin{abstract}
\section*{Abstract}
  Recent advances in deep learning have improved the performance of many Natural Language Processing (NLP) tasks such as translation, question-answering, and text classification. However, this improvement comes at the expense of model explainability. Black-box models make it difficult to understand the internals of a system and the process it takes to arrive at an output. Numerical (LIME, Shapley) and visualization (saliency heatmap) explainability techniques are helpful; however, they are insufficient because they require specialized knowledge. These factors led rationalization to emerge as a more accessible explainable technique in NLP. Rationalization justifies a model's output by providing a natural language explanation (rationale). Recent improvements in natural language generation have made rationalization an attractive technique because it is intuitive, human-comprehensible, and accessible to non-technical users. Since rationalization is a relatively new field, it is disorganized. As the first survey, rationalization literature in NLP from 2007-2022 is analyzed. This survey presents available methods, explainable evaluations, code, and datasets used across various NLP tasks that use rationalization. Further, a new subfield in Explainable AI (XAI), namely, Rational AI (RAI), is introduced to advance the current state of rationalization. A discussion on observed insights, challenges, and future directions is provided to point to promising research opportunities.
\end{abstract}



\keywords{Rationalization, Rational AI, Natural Language Generation, Natural Language Processing, Language Modeling, Explainable AI}

\maketitle
\section{Introduction}
\label{section:intro}
The commercialization of NLP has grown significantly in the past decade. Text has a ubiquitous nature which enables many practical NLP use cases and applications, including but not limited to text classification, fact-checking, machine translation, text2speech, and others, which significantly impact our society. Despite its diverse and practical applications, NLP faces many challenges; an important one is explainability \cite{Intro7}.

In the past, NLP systems have traditionally relied on white-box techniques. These techniques - rules, decision trees, hidden Markov models, and logistic regression - are inherently explainable \cite{Intro4}. The recent developments in deep learning have contributed to the emergence of black-box architectures that improve task performance at the expense of model explainability. Such black-box predictions make understanding how a model arrives at a decision challenging. This lack of explainability is a significant cause of concern for critical applications. For example, directly applying natural language generation methods to automatically generate radiology reports from chest X-ray images only guarantees that the produced reports will look natural rather than contain correct anatomically-aware information \cite{liu19a}. 

Similarly, Visual Question Answering (VQA) systems are known to learn heavy language priors \cite{AgrawalBP16}. Thus, a lack of transparency can affect the decision-making process and may lead to the erosion of trust between humans and Artificial Intelligence (AI) systems. This can further jeopardize users' safety, ethics, and accountability if such a system is deployed publicly \cite{Intro7}. Considering the utilization of NLP in healthcare, finance, and law domains, all of which can directly affect human lives, it can be dangerous to blindly follow machine predictions without fully understanding them. For instance, a physician following a medical recommendation or an operation procedure for a patient without full knowledge of the system can do more harm than good. In addition, systems employing Machine Learning (ML), such as most current NLP methods, are prone to adversarial attacks where small, carefully crafted local perturbations can maximally alter model predictions, essentially misguiding the model to predict incorrectly but with high confidence \cite{Intro8}. The bar for the ethical standards and the accountability required to maintain NLP systems continue to increase as these systems become more opaque with increasingly complex networks and algorithms.

There has been significant research focus on enabling models to be more interpretable, i.e., allowing humans to understand the internals of a model \cite{Intro10}. However, due to the lack of completeness, interpretability alone is not enough for humans to trust black-box models. \textit{Completeness} is the ability to accurately describe the operations of a system that allows humans to anticipate its behavior better. \citet{Intro10} argue that \textit{explainability} improves on interpretability as a technique to describe the model's decision-making process of arriving at a prediction and the ability to be verified and audited. Therefore, models with explainability are interpretable and complete. In this survey, the focus is on explainability and mainly on the \textbf{outcome explanation problem} where \citet{Intro2} describe explainability as ``the perspective of an end-user whose goal is to understand how a model arrives at its result''. 

In NLP, there exist various explainable techniques such as LIME (Local Interpretable Model-Agnostic Explanations) \cite{SG17}, Integrated Gradients \cite{IntegratedGradients}, and SHAP (Shapley Additive Explanations) \cite{SHAP}. Despite the availability of these methods, many require specialized knowledge to understand their underlying processes, which makes them indecipherable and inaccessible for the general users or audience, which we refer to as the nonexperts, hence limiting usability. In the following sections, we share how rationalization addresses these problems and helps improve explainability for nonexpert users.

The structure of this literature survey is as follows. In Section \ref{sec:background}, we share the background and intuition for rationalization. In Section \ref{sec:method}, we explain our paper collection methodology. We identify and describe the most commonly used rationalization techniques and point to available papers adopting them in Section \ref{sec:ratTech}. In Section \ref{sec:extAbs}, we compare and contrast abstractive and extractive rationalization techniques. We conclude with a discussion of the open challenges and promising future directions in Section \ref{sec:discuss}.

\section{Background}
\label{sec:background}

One of the emerging explainable techniques for NLP applications is rationalization \cite{AK25}. \textbf{Rationalization} provides explanations in natural language to justify a model's prediction. These explanations are rationales, which present the input features influencing the model's prediction. The reasoning behind the prediction could be understood simply by reading the explanation/rationale, thereby revealing the model's decision-making process. Rationalization can be an attractive technique because it is human-comprehensible and allows individuals without domain knowledge to understand how a model arrived at a prediction. It essentially allows the model to "talk for themselves" \cite{Intro9, ElsevierSub2}. This technique is a part of a subset of explainability because it enables models to be interpretable and complete, as shown in Figure \ref{fig:subsets}. Specifically, rationalization provides a local explanation since each prediction has a unique explanation rather than one for the entire model. Local explanations can be categorized into two groups: local post-hoc and local self-explaining. \citet{Intro4} present local post-hoc methods as explaining a single prediction after the model predicts and local self-explaining methods as simultaneously explaining and predicting. 

\begin{center}
\includegraphics[scale=0.3]{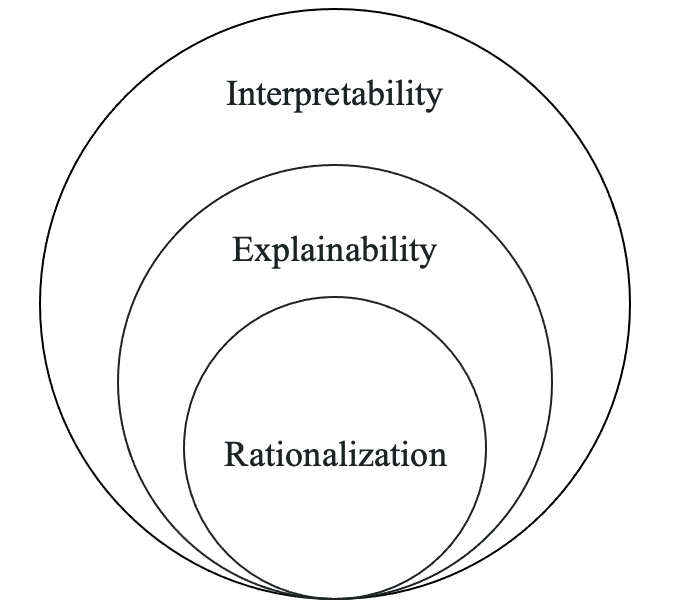}
\captionof{figure}{The Rationalization Field}
\label{fig:subsets}
\end{center}

Rationalization in NLP was first introduced in 2007 \cite{SG6}. As described in Section 4.6, the objective was to use annotator rationales to improve task performance for text categorization. Interestingly, explainability was not the core objective. However, explainability is an advantage of rationalization because it makes the model inherently explainable even if used in the context of task improvement \cite{exclaim}.

\begin{figure}[h]
    \centering
    \subfloat[\centering Extractive Rationalization \cite{SG22}]{{\includegraphics[width=0.45\columnwidth]{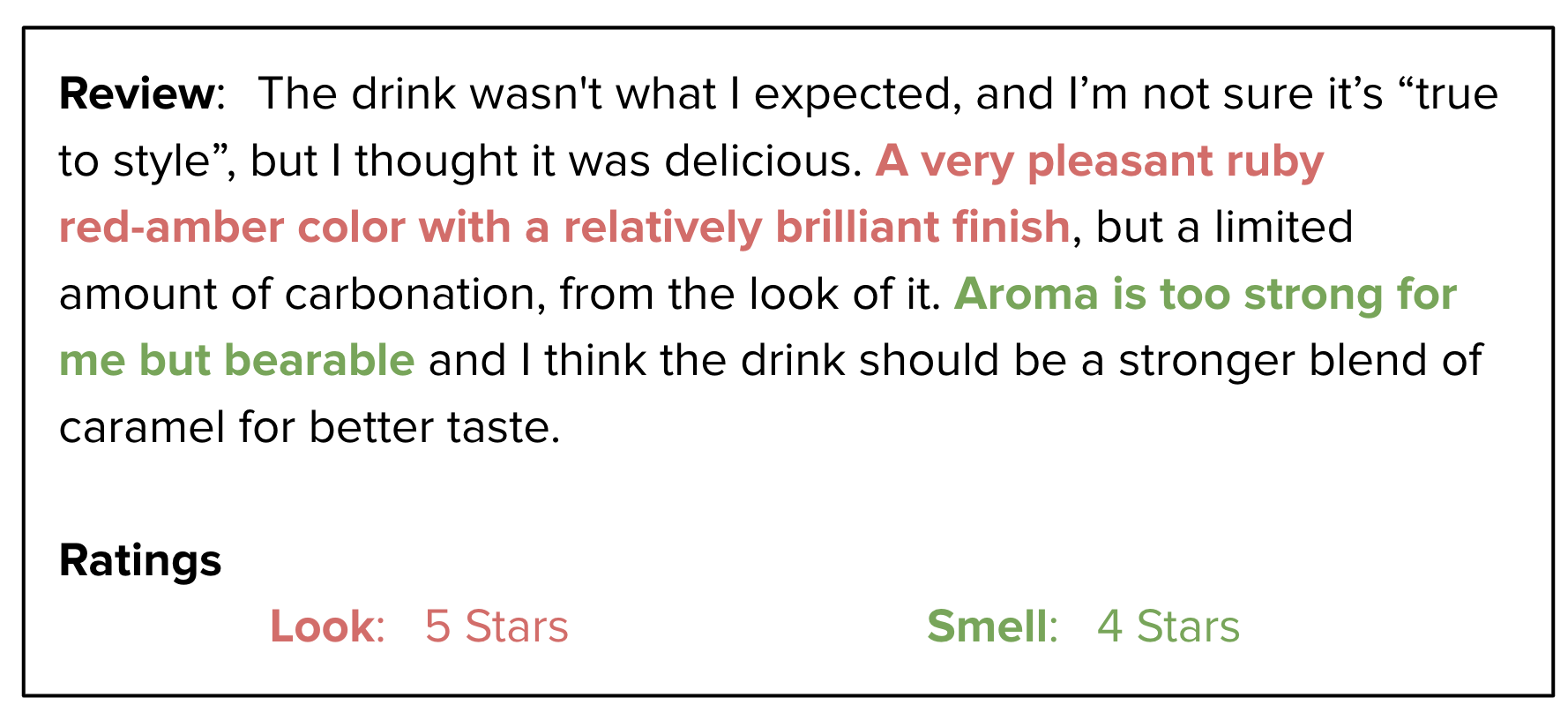} }}%
    \qquad
    \subfloat[\centering Abstractive Rationalization \cite{AK9}]{{\includegraphics[width=0.45\columnwidth]{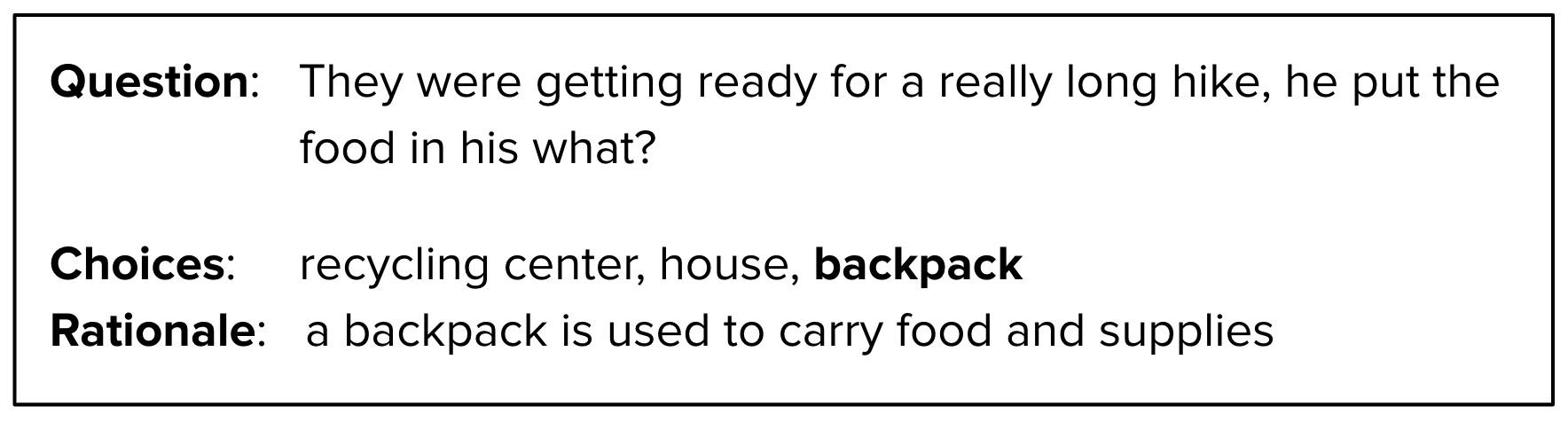} }}%
    \caption{Rationalization Types}
    \label{fig:ratTypes}%
\end{figure}

Our literature review found that rationalization can be further divided into two major groups: abstractive and extractive \cite{Disc1}. In extractive rationalization, important features or sentences from the input data are extracted as rationales to support the prediction. In contrast, abstractive rationalization is a generative task in which novel sentences are generated using new words or paraphrasing existing sentences. This is typically accomplished through the use of a language model such as T5 \cite{T5Paper} or GPT (Generative Pre-trained Transformer) \cite{GPTPaper}. Figure \ref{fig:ratTypes} demonstrates the usage of the two explanation types with examples. 


Recent advances in explainable NLP have led to a significant increase in rationalization research. Further, at present, the field of rationalization is disorganized. Thus, the motivations for this survey are — a) formally define rationalization, b) present and categorize the well-cited techniques based on NLP tasks, and c) discuss current trends and future insights on the field. Thus, our primary goal is to provide future researchers with a comprehensive understanding of the previously scattered state of rationalization. The key contributions of this paper are as follows.

\begin{enumerate}
\item First literature to survey the field of rationalization in NLP.
\item Introduction of a new subfield called Rational AI (RAI) within Explainable AI (XAI).
\item A comprehensive list of details on available rationalization models, XAI evaluations, datasets, and code are provided to guide future researchers.
\item Presents NLP Assurance as an important method for developing more trustworthy and reliable NLP systems.
\end{enumerate}

\subsection{Related Surveys}

\begin{table*}[h]
  \caption{Related Survey Papers}
  \label{tab:relatedSurveys}
    \resizebox{\textwidth}{!}{
  \begin{tabular}{ccccc}
    \toprule
    No. & Survey Title & Year & Papers Reviewed\\
    \midrule
    1 & A Survey on the State of Explainable AI for Natural Language Processing \cite{Intro4} & 2020 & 50\\
    2 & A Survey on Explainability in Machine Reading Comprehension \cite{Intro5} & 2020 & 69\\
    3 & Post-hoc Interpretability for Neural NLP: A Survey \cite{Intro7} & 2021 & 27\\
    \bottomrule
  \end{tabular}}
\end{table*}

\citet{Intro4} note that previous surveys in XAI are broadly focused on AI without a specific narrow domain focus. Their work primarily focuses on surrogate-based explainability methods. NLP publications in recent years further demonstrate that this distinction is less relevant and valuable in the NLP domain because “the same neural network can be used not only to make predictions but also to derive explanations.” Therefore, surveying the field of Explainable AI (XAI) in NLP requires NLP-specific methods that are different from the standard XAI methods that are widely known.

\citet{Intro5} survey the use of explanations specifically in Machine Reading Comprehension (MRC). The authors describe MRC papers that support explanations and provide a detailed overview of available benchmarks. Further, \citet{Intro7} briefly discuss rationalization and natural language explanations using a question and answer approach, CAGE (Commonsense Auto-Generated Explanations), \cite{AK9} as an example. Thus, this raises the question - \textit{how can rationalization be generalized and applied to other tasks in the NLP domain?} However, until now, no comprehensive literature review on rationalization has been available for the prominent NLP tasks. Thus, through this survey paper, we attempt to address this need. 

\subsection{Definitions}
In order to provide clarity and to distinguish terms that are typically used interchangeably in published literature, the following definitions are provided, which are used throughout the paper.

\vspace{0.5cm}

\noindent \textbf{Black-box Model}: A “machine-learning obscure model, whose [architecture] internals are either unknown to the observer or they are known but uninterpretable by humans” \cite{Intro2}.

\vspace{0.5cm}

\noindent \textbf{Interpretability}: Interepretability “aims at developing tools to understand and investigate the behavior of an AI system” \cite{Intro5}. \citet{Intro3} augments this definition by adding that interpretability tools allow us to “explain or to present in understandable terms to a human” what the AI system is performing.

\vspace{0.5cm}

\noindent \textbf{Explainability}: There is no consensus on the nature of explanations since they are entirely task-dependent and AI embraces a wide variety of tasks \cite{Intro6}. We treat explainability as a specialization of interpretability where the aim of explainability is to design inherently interpretable models, capable of performing transparent inference through the generation of an explanation for the final prediction \cite{Intro5}. In this survey, we focus on black-box explainability, specifically, as the outcome explanation problem. 

\vspace{0.5cm}

\noindent \textbf{Rationalization}: The term rationalization is interchangeable with explanation or justification. Rationalization has rarely been formally defined in the context of NLP, therefore, we propose the following definition.

\vspace{0.2cm}

\noindent\textit{Rationalization justifies a model's output by providing a natural language explanation. This is accomplished by either extracting text fragments from the input (extractive rationalization) or by generating a novel explanation (abstractive rationalization).
}

\vspace{0.5cm}


 


\noindent\textbf{NLP Assurance}: A  process that is applied at all stages of the NLP development lifecycle to ensure that all outcomes are valid, verified, trustworthy, and explainable to a nonexpert, ethical in the context of its deployment, unbiased in its learning, and fair to its users. This definition is adopted from \citet{Intro1} and modified to fit the scope of the NLP domain. 

\section{Methodology}
\label{sec:method}

The following are the inclusion-exclusion criteria for our publications collection methodology. The first known use of rationalization was in the year 2007. Our survey focuses on the domain of Natural Language Processing from 2007 to early 2022 \cite{SG6}. We have included peer-reviewed publications within this range that include a significant rationalization component as a method to provide explainability. We defined \textit{significance} as rationalization being the main component of their research methodology and approach. We have eliminated a number of publications that are either not entirely in the NLP domain or do not contain a significant rationalization component. 

For identifying and selecting articles, the range of keywords and topics was limited to the following in the NLP domain:  rationalization, explanation, justification, and explainable NLP. Thus, this survey includes reviews of the articles from journals, books, industry research, dissertations, and conference proceedings from commonplace AI/NLP venues such as ACL, EMNLP, NAACL, AAAI, NeurIPS, and others. Finally, these articles are categorized by important NLP tasks, as shown in Table \ref{tab:nlpTasks}. In recent years, there has been an increase in focus on explainability in NLP after a rise in deep learning techniques \cite{Intro4}. Due to this, a majority of the papers collected were from recent years (2016 and onwards), as illustrated in Figure \ref{fig:yearlypapers}. 

\begin{table*}[h]
  \caption{NLP Tasks Surveyed}
  \label{tab:nlpTasks}
  \resizebox{\textwidth}{!}{
  \begin{tabular}{cc}
    \toprule
    NLP Task & Definition \\
    \midrule
   Machine Reading
Comprehension (MRC) & Enabling a model to answer questions regarding a given context \cite{AK1}.  \\
 Commonsense Reasoning &    Going beyond pattern recognition to make inferences using world knowledge \cite{AK7, AK8}.     \\
  Natural Language Inference &    Determining if a hypothesis entails or contradicts a premise \cite{AK12}.     \\
   Fact Checking &    Classifying if a claim is either true or false based on evidence \cite{AK20, exclaim}.     \\
  Sentiment Analysis &   Quantifying whether the textual data has a positive, negative, or neutral emotion \cite{SG10}.     \\
  Text Classification & Categorizing textual data by automatically assigning labels \cite{SG5}.\\
  Neural Machine Translation & Translating languages using deep neural networks \cite{SG2}. \\
    \bottomrule
  \end{tabular}
  }
\end{table*}


The availability of relevant articles was limited. After following the above approach, 33 articles were downselected to be the primary focus of this paper's discussion. Instead of providing a broad and surface-level understanding of the work, we focus on sharing in-depth the most important approaches and progress made in each NLP task. Overall, we selected six articles in multiple NLP domains, five on Machine Reading Comprehension and Sentiment Analysis, four on Text Classification, Fact-Checking and Commonsense Reasoning, and three on Natural Languages Inference, and two articles on Neural Machine Translation (NMT).

\begin{center}
\includegraphics[scale=0.43]{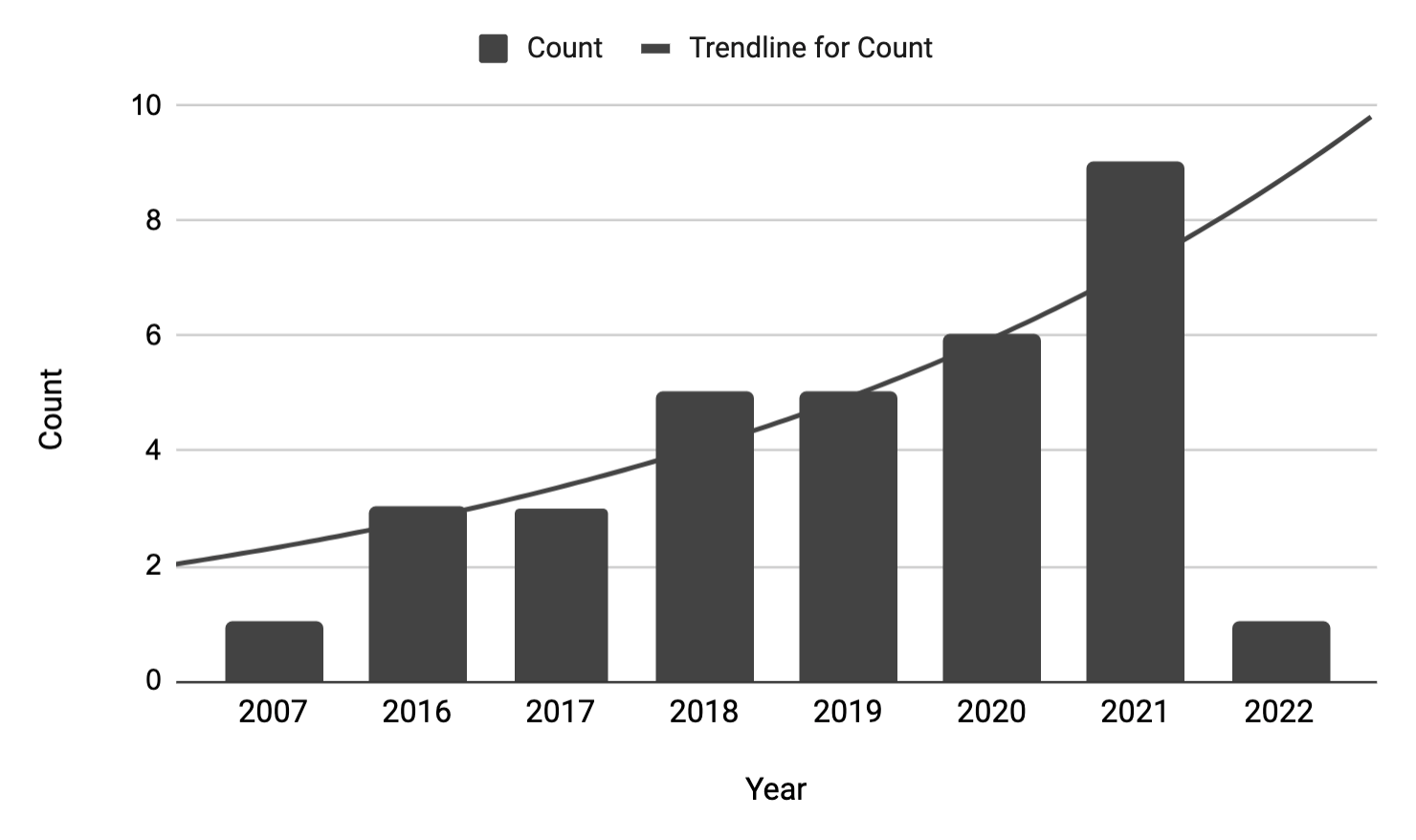}
\captionof{figure}{Collected Papers Per Year}
\label{fig:yearlypapers}
\end{center}

\section{Rationalization Techniques}
\label{sec:ratTech}

In this section, we discuss relevant papers and their rationalization techniques categorized by the NLP tasks listed in Figure \ref{tab:nlpTasks}. Tables with important information on the papers for each subsection are presented at the beginning.

\subsection{Machine Reading Comprehension}

\begin{table*}[h]
  \caption{Machine Reading Comprehension Papers}
  \label{tab:mrctable}
    \resizebox{\textwidth}{!}{
  \begin{tabular}{cccccccc}
    \toprule
    Paper & Name & Year  & Explanation & Models & XAI Metric & Dataset & Code\\
    \midrule
    \cite{MRC1} & - & 2017 & Extractive & TF-IDF, FFNN & - & AI2 Science, Aristo Mini & -\\
    \cite{MRC2} & - & 2017 & Extractive & LSTM, Seq2Seq & - & AQuA & \checkmark \\
    \cite{AK2} & OpenBookQA & 2018 & Abstractive & BiLSTM Max-out & - & OpenBookQA,  & \checkmark\\
    \cite{AK3} & WorldTree V2 & 2018 & Abstractive & TF-IDF, BERT & - & WorldTree V2 & \checkmark\\
    \cite{AK4} & FiD-Ex & 2021 & Extractive & T5, BERT-to-BERT & - & Natural Questions & -\\
   
    \bottomrule
  \end{tabular}}
\end{table*}

MRC enables a model to answer questions regarding a given context \cite{AK1}. For this reason, it also frequently referred to as Question Answering (QA) Systems. For MRC applications, we found five recent articles from which three articles provide novel datasets \cite{AK2}, \cite{AK3}, \cite{MRC2} and the remaining articles \cite{AK4} and \cite{MRC1} each propose a new MRC framework.
 
The first article, published in 2018, presented a new question-answering dataset based on the open book exam environment for elementary-level science - OpenBookQA \cite{AK2}. This dataset consists of two components – i) Questions (Q): a set of 5,958 multiple choice questions and ii) Facts (F): a set of 1,326 diverse facts about elementary level science. This dataset was further tested for evaluating the performance of existing QA systems and then compared with the human performance. The results indicated that human performance was close to 92\%, but many existing QA systems showed poor performance close to the random guessing baseline of 25\%. Additionally, the authors found that simple neural networks achieved an accuracy of about 50\%, but it is still not close to the human performance, about 92\%. Recently an extension of the WorldTree project \cite{AK5}, i.e., WorldTree V2 \cite{AK3}, is presented. The main goal of this project is to generate a science domain explanation with a supporting semi-structured knowledge base. The WorldTree project is a part of explainable question answering tasks that provide answers to natural language questions and their human-readable explanations for why the answers are correct. \citet{AK3} notes that most multi-hop inference models could not demonstrate combining more than two or three facts to perform inference. However, here the authors merge, on average, six facts from a semi-structured knowledge base of 9216 facts. Thus, this resulted in the WorldTree V2 corpus for standardized science questions. This corpus consists of 5100 detailed explanations to support training and instrumenting multi-hop inference question answering systems. 
 
\citet{AK4} demonstrates a new MRC framework called FiD-Ex (Extractive Fusion-in-Decoder). It has been noted that seq2seq (Sequence to Sequence) models work well at generating explanations and predictions together. However, these models require a large-labeled dataset for training and bring a host of challenges such as fabricating explanations for incorrect predictions, the difficulty of adapting to long input documents, etc. Thus, to tackle these challenges, the Fid-Ex framework includes sentence markers to encourage extractive explanations and intermediate fine-tuning for improving few-shot performance on open-domain QA datasets. This new framework is tested on ERASER (Evaluating Rationales And Simple English Reasoning) datasets and their benchmarks for evaluations \cite{SG16}. This experiment concludes that FiD-Ex significantly improves upon prior work on the explanation metrics and task accuracy on supervised and few-shot settings.

\citet{MRC1} proposes a new neural network architecture that re-ranks answer justifications as an intermediate step in answer selection. This new approach alternates between a max pooling layer and a shallow
neural network (with ten nodes, glorot uniform initializations, tanh activation, and L2
regularization of 0.1) for providing a justification. This approach contains three
components: 1) retrieval component, which retrieves a pool of candidates' answer justification,
2) extractor, which extracts the features and 3) scores, which perform the scoring of the answer
candidate based on the pool of justifications. The authors used 8th-grade science questions
provided by Allen Institute for Artificial Intelligence (AI2) for evaluations. The training set
includes 2500 questions with four options, and the test set consists of 800 publicly released
questions. Further, a pool of candidate justifications corpora containing 700k sentences from
StudyStack and 25k sentences from Quizlet is used. The top 50 sentences were retrieved as a
set of candidate justification. For model tuning, the authors used five-fold cross-validation,
and during testing, the model architecture and hyperparameters were frozen.
The authors compared results using two baselines: IR baseline and IR++. They concluded that this new approach showed better accuracy and justification
quality while maintaining near state-of-the-art performance for the answer selection task.

\citet{MRC2} presented a dataset and an approach that provides answer rationales,
sequences of natural language, and human-readable mathematical expressions for solving
algebraic word problems. The authors proposed a sequence-to-sequence model which
generates a sequence of instructions and provides the rationales after selecting the answer.
For this purpose, a two-layer LSTM (Long Short-Term Memory) with a hidden size of 200 and word embedding with a size of
200 is utilized. Further, the authors also built a dataset containing 100,000 problems in which
each question is decomposed into four parts – two inputs and two outputs. This new dataset is
used for generating rationales for math problems and for understanding the quality of
rationales as well as the ability to obtain a correct answer. Further, the authors used an
attention-based sequence to sequence model as a baseline and compared results based on
average sentence level perplexity and BLEU-4 (Bilingual Evaluation Understudy). The authors noted that this new approach could
outperform the existing neural models in the ability to solve problems and the fluency of the
generated rationales.

\subsection{Commonsense Reasoning}

\begin{table*}[h]
  \caption{Commonsense Reasoning Papers}
  \label{tab:crtable}
  \resizebox{\textwidth}{!}{
  \begin{tabular}{cccccccc}
    \toprule
    Paper & Name & Year  & Explanation & Models & XAI Metric & Dataset & Code\\
    \midrule
    \cite{SG23} &    -   & 2018 & Extractive  & LSTM, Seq2Seq              &  - & -  & -  \\
    \cite{AK9} & CAGE  & 2019 & Abstractive & GPT, BERT                               & - & CoS-E, CommonsenseQA                & \checkmark \\
\cite{AK10} & RExC  & 2021 & Extractive  & Transformer                             & - & ComVE, e-SNLI, COSe, e-SNLI-VE, VCR & -  \\
\cite{AK11} & DMVCR & 2021 & Extractive  & LSTM, BERT & - & VCR                                 & \checkmark \\
    \bottomrule
  \end{tabular}}
\end{table*}

Commonsense knowledge helps humans navigate everyday situations. Similarly, commonsense reasoning in NLP is the ability for a model to go beyond pattern recognition and use world knowledge to make inferences \cite{AK7, AK8}. On commonsense reasoning, we found four articles, and all of them provide unique solutions that contribute to the development of commonsense reasoning frameworks.
 
\citet{AK8} demonstrates a solution for commonsense reasoning using LSTM encoder and decoder. The main goal was to convert the actions of an autonomous agent into natural language using neural machine translation. For this purpose, the authors built a corpus of thoughts of people as they complete tasks in the Frogger game which are then stored as states and actions. In the next step, LSTM encoder and decoder are used to translate actions as well as states into natural language. Lastly, the authors used the BLEU score to calculate sentence similarity and assessed the accuracy for selecting the best rationale. The authors also conducted a survey to evaluate the rationales based on human satisfaction. The Frogger experiment is concluded with Encoder-Decoder framework outperforming the baselines and demonstrates that the use of game theory approaches for generating rationales is a promising technique. Similar work from \cite{ElsevierSub3, ElsevierSub4, ElsevierSub5, ElsevierSub6} has further advanced progress in this space.

Further, it is noted that deep learning model performance is poor when used in tasks that require commonsense reasoning due to limitations with available datasets. To tackle this problem, \citet{AK9} developed the Commonsense Auto-Generated Explanations (CAGE) framework for generating explanations for Commonsense Question Answering (CQA). The authors also created a new dataset – Common Sense Explanations (CoS-E) – by collecting human explanations for commonsense reasoning and highlighting annotations. From this paper, the authors concluded that CAGE could be effectively used with pre-trained language models to increase commonsense reasoning performance.
 
Recently, \citet{AK10} and \citet{AK11} presented novel solutions for commonsense reasoning. \citet{AK10} focused on the Natural Language Expiations (NLEs), which are more detailed than Extractive rationales but fall short in terms of commonsense knowledge. In this solution, the authors proposed a self-rationalizing framework RExC (Rationales, Explanations, and Commonsense). RExC first extracts rationales that act as features for the prediction then expands the extractive rationales using commonsense resources. In the last step, RExC selects the best suitable commonsense knowledge for generating NLEs and a final prediction. The authors tested RExC on five tasks - three natural language understanding tasks and two vision language understanding tasks. Overall, the results indicated improvement in the quality of extractive rationales and NLEs that bridges the gap between task performance and explainability. On the other hand, Tag et al. \cite{AK11} focused on Visual Commonsense Reasoning (VCR).  They focused on a problem when a question with a corresponding input image is given to the system, and it attempts to predict an answer with a rationale statement as the justification. To explore this, author presented a multi-model approach by combining Computer Vision (CV) and NLP. Their approach leverages BERT and ResNet50 (Residual neural network) as the feature representation layer and BiLSTM (Bidirectional LSTM) and Attention for the multimodal feature fusion layer. These layers are then concatenated into an LSTM network for the encoder layer before passing into the classifier for the prediction layer. This was tested on the benchmark VCR dataset and it indicated significant improvements over existing methods and it also provided a more interpretable intuition into visual commonsense reasoning.

\subsection{Natural Language Inference}

\begin{table*}[h]
  \caption{Natural Language Inference Papers}
  \label{tab:crtable}
  \resizebox{\textwidth}{!}{
  \begin{tabular}{cccccccc}
    \toprule
    Paper & Name & Year  & Explanation & Models & XAI Metric & Dataset & Code\\
    \midrule
   \cite{AK13} & e-SNLI & 2018 & Abstractive & BiLSTM, Seq2Seq  & - & e-SNLI      & \checkmark \\
\cite{AK15} & NILE   & 2020 & Abstractive & GPT-2, RoBERTa & - & e-SNLI      & \checkmark \\
\cite{AK16} &   -     & 2020 & Abstractive & T5                          & - & CoS-E, SNLI & \checkmark \\
    \bottomrule
  \end{tabular}}
\end{table*}

Natural Language Inference (NLI) task helps with identifying a natural language hypothesis from a natural language premise \cite{AK12}. For this application, we found three articles. The first article presents a new dataset - e-SNL (explanation-augmented Stanford Natural Language Inference) \cite{AK13} – and the other two articles discuss approaches that can improve NLI. 
 
Camburu et al. \cite{AK13} extended the Stanford NLI (SNLI) \cite{AK14} dataset by providing human-annotated explanations for the entailment relations. This new dataset – e-SNLI - is used in a series of classification experiments involving LSTM-based networks for understanding its usefulness for providing human-interpretable full-sentence explanations. The authors also evaluated these explanations as an additional training signal for improving sentence representation and transfer capabilities of out-of-domain NLI datasets. Thus, from these experiments, the authors conclude that e-SNLI can be used for various goals mentioned above and also be utilized for improving models as well as asserting their trust.
 
Another issue with NLI is the faithfulness of the generated explanations, tackled by Kumar \& Talukdar \cite{AK15} and Wiegreffe et al. \cite{AK16}. Kumar \& Talukdar \cite{AK15} mentioned that existing methods do not provide a solution for understanding correlations of the explanations with the model’s decision-making and this can affect the faithfulness of the generated explanations. Considering this problem, the authors proposed and presented a new framework – NILE (Natural language Inference over Label-specific Explanations). The NILE framework can generate natural language explanations for each possible decision and process these explanations to produce a final decision for the classification problems. To test this approach, the authors used two datasets - SNLI and e-SNLI – and compared NILE with baseline and other existing approaches based on explanation accuracy, in-domain evolution sets (SNLI), and on out-of-domain examples (train on SNLI and test on MNLI \cite{AK17}). Based on the first 100 SNLI test samples, the results indicated that NILE variants are comparable with the ETPA (Explain Then Predict Attention) baseline, and NILE explanations generalize significantly better on out-of-domain examples. For out-of-domain examples (MNLI), results showed that the percentage of correct explanations in the subset of correct label predictions was significantly better for all the NILE variants. Thus, the authors concluded that NILE is an effective approach for accurately providing both labels and explanations. Further, Wiegreffe et al. \cite{AK15} also focused on the need for faithfulness for denoting the model’s decision-making process by investigating abstractive rationales. The author proposed two measurements - robustness equivalence and feature importance agreement – to investigate the association of the labels and predicted rationales, which are required for a faithful explanation. This investigation was performed on CommonsenseQA \cite{AK18} and SNLI dataset using T5-based models \cite{AK19}. The results indicated that state-of-the-art T5-based join models demonstrate desirable properties and potential for producing faithful abstractive rationales.

\subsection{Fact-Checking}

\begin{table*}[h]
  \caption{Fact-Checking Papers}
  \label{tab:crtable}
  \resizebox{\textwidth}{!}{
  \begin{tabular}{cccccccc}
    \toprule
    Paper & Name & Year  & Explanation & Models & XAI Metric & Dataset & Code\\
    \midrule
   \cite{AK23}  & LIAR-PLUS & 2018 & Extractive & SVM, BiLSTM & - & LIAR-PLUS & \checkmark \\
\cite{AK24}  &     -      & 2019 & Extractive & BERT                             & - & FEVER     & \checkmark \\
\cite{AK25} &      -     & 2020 & Extractive & DistilBERT                       & - & LIAR-PLUS &  - \\
\cite{AK26}  & RERRFACT  & 2022 & Extractive & RoBERTa, BioBERT         & - & SCIFACT   & - \\
    \bottomrule
  \end{tabular}}
\end{table*}

Fact-checking has become a popular application of NLP in recent years given its impact on assisting with misinformation and a majority of the work has been with claim verification \cite{AK20, exclaim}. Based on a paper published in 2016, there are 113 active fact-checking groups and 90 of which were established after 2010 \cite{AK21}. This indicates the growth of the fact-checking application. Considering the scope of this literature review, we found four articles on fact-checking. Two of the studies in this section present novel datasets, and the remaining two provide new techniques to improve fact-checking.
 
In 2017, a large dataset for the fact-checking community called LIAR \cite{AK22} was introduced, including POLITIFACT data. Most works on this data were focused on using the claim and its speaker-related metadata to classify whether a verdict is true or false. The evidence - an integral part of any fact-checking process - was not part of the LIAR and was overlooked. Thus, in 2018 Tariq et al. \cite{AK23} extended the LIAR dataset to LIAR-plus by including the evidence/justification. The authors treated the justification as a rationale for supporting and explaining the verdict. Further, they used Feature-based Machine Learning models (Logistic Regression and Support Vector Machine) and deep learning models (Bi-Directional Long Short-term Memory (BiLSTM) and Parallel-BiLSTM) for binary classification tasks to test the data. The results demonstrated a significant performance improvement in using the justification in conjunction with the claims and metadata. Further, Hanselowsk et al. \cite{AK24} introduced a new corpus for training machine learning models for automated fact-checking. This new corpus is based on different sources (blogs, social media, news, etc.) and includes two granularity levels - the sources of the evidence and the stance of the evidence towards the claim - for claim identification. Authors then used this corpus to perform stance detection, evidence extraction and claim validation experiments. In these experiments, a combination of LSTMs, baseline NN, pre-trained models have been used, and their results are compared based on precision, recall, and F1 macro. The results indicated that fact-checking using heterogeneous data is challenging to classify claims correctly. Further, the author claims that the fact-checking problem defined by this new corpus is more difficult compared to other datasets and needs more elaborate approaches to achieve higher performance.   
 
It has been noted that the fact-checking systems need appropriate explainability for the verdicts they predict. The justifications that are human-written can help to support and provide context for the verdicts, but they are tedious, unscalable, and expensive to produce \cite{AK25}. Considering this issue, Atanasova et al. \cite{AK25} proposed that the creation of the justifications needs to be automated to utilize them in a large-scale fact-checking system. The authors presented a novel method that automatically generates the justification from the claim’s context and jointly models with veracity prediction. Further, this new method is then tested on the LIAR dataset \cite{AK22} for generating veracity explanations. The results indicated that this new method could combine predictions with veracity explanations, and manual evaluations reflected the improvement in the coverage and quality of the explanations. Another important domain in which fact-checking is useful is Science. Researching and providing substantial evidence to support or refute a scientific claim is not a straightforward task. It has been seen that scientific claim verification requires in-depth domain expertise along with tedious manual labor from experts to evaluate the credibility of a scientific claim. Considering this problem, Rana et al. \cite{AK26} proposed a new framework called RERRFACT (Reduced Evidence Retrieval Stage Representation) for classifying scientific claims by retrieving relevant abstracts and training a rationale-selection model. RERRFACT includes a two-step stance prediction that differentiates non-relevant rationales then identifies a claim's supporting and refuting rationales. This framework was tested on the SCI-FACT dataset \cite{AK27} and performed competitively against other language model benchmarks on the dataset leaderboard.

\subsection{Sentiment Analysis}

\begin{table*}[h]
  \caption{Sentiment Analysis Papers}
  \label{tab:crtable}
  \resizebox{\textwidth}{!}{
  \begin{tabular}{cccccccc}
    \toprule
    Paper & Name & Year  & Explanation & Models & XAI Metric & Dataset & Code\\
    \midrule
   \cite{SG22}  &  -      & 2016 & Extractive & LSTM, RCNN           & - & BeerAdvocate, AskUbuntu     & \checkmark \\
\cite{SG11}  & CREX   & 2019 & Extractive & CNN, LSTM & - & BeerAdvocate, MovieReview   & -  \\
\cite{SG12} &   -     & 2019 & Extractive & RA-CNN, AT-CNN            & - & MovieReview                 &  - \\
\cite{SG13}  & A2R    & 2021 & Extractive & BiGRU                     & - & BeerAdvocate, MovieReview   & \checkmark \\
\cite{SG14}  & ConRAT & 2021 & Extractive & CNN, BiGRU                & - & AmazonReviews, BeerAdvocate & -  \\
    \bottomrule
  \end{tabular}}
\end{table*}

Sentiment Analysis is a subset of the text classification field \cite{SG5}. It focuses specifically on the "computational study of people’s opinions, sentiments, emotions, appraisals, and attitudes towards entities such as products, services, organizations, individuals, issues, events, topics and their attributes” \cite{SG10}. The use of rationales to support sentiment analysis models in NLP is widely used compared to other NLP tasks. We identified five papers in this field.

In 2016, Lei et al. \cite{SG22} pioneered rationalization in sentiment analysis by proposing a problem: “prediction without justification has limited applicability”. To make NLP outcomes more transparent, the authors propose an approach to extract input text which serves as justifications or rationales for a prediction. These are fragments from the input text which themselves are sufficient to make the same prediction. Their implementation approach includes a generator and an encoder architecture. The generator determines which can be potential candidates for a rationale from the input text. Those candidates are fed into the encoder to determine the prediction and the rationales are not provided during training. They employ an RCNN (Region-based Convolutional Neural Network) and an LSTM architecture and when compared with each other the RCNN performed better. The experiment was conducted on the BeerAdvocate dataset. The paper's approach outperforms attention-based baseline models. They also demonstrate their approach on a Q\&A retrieval task indicating that leveraging rationales for sentiment analysis tasks is very beneficial.

Similarly, \citet{SG11} claim that explainability alone is not sufficient for a DNN (Deep Neural Network) to be viewed as credible unless the explanations align with established domain knowledge.” In essence, only the correct evidences are to be used by the networks to justify predictions. In this paper, the authors define credible DNNs as models that provide explanations consistent with established knowledge. Their strategy is to use domain knowledge to improve DNNs credibility. The authors explore a specific type of domain knowledge called a rationale which are the salient features of the data. They propose an approach called CREX (Credible Explanation), which regularizes DNNs to use the appropriate evidence when making a decision for improved credibility and generalization capability. During training, instances are coupled with expert rationales and the DNN model is required to generate local explanations that conform to the rationales. They demonstrate it on three types of DNNs (CNN, LSTM, and self-attention model) and various datasets for testing. Results show that the CREX approach allows DNNs to look at the correct evidences rather than the specific bias in training dataset. Interestingly, they point that incorporating human knowledge does not always improve neural network performance unless the knowledge is very high quality.

Many papers published in the rationalization field indicate that a machine learning system learning with human provided explanations or “rationales” can improve its predictive accuracy \cite{SG6}. \citet{SG12} claim that this work hasn’t been connected to the XAI field where machines attempt to explain their reasoning to humans. The authors attempt to show in their paper that rationales can improve machine explanations as evaluated by human judges. Although automated evaluation works, \citet{SG12} believe that since the explanations are for users, therefore humans should directly evaluate them. The experiment is done by using the movie reviews dataset and by having a supervised and an unsupervised CNN model for a text classification task. They use attention mechanism and treat the rationales as supervision in one of the CNN models. Results indicate that a supervised model trained on human rationales outperforms the unsupervised on predictions. The unsupervised is the model where the rationales/explanations are learned without any human annotations.

The selective rationalization mechanism is commonly used in complex neural networks which consist of two components – rationale generator and a predictor. This approach has a problem of model interlocking which arises when the predictor overfits to the features selected by the generator. To tackle this problem this paper proposes a new framework A2R which introduces a third component for soft attention into the architecture \cite{SG13}. The authors have used BeerAdvocate and MovieReview for understanding the effectiveness of the framework. The authors compared results from A2R with the original rationalization technique RNP (Rationalizing Neural Predictions) along with 3PLAYER, HARD-KUMA and BERT-RNP. For implementation authors have used bidirectional Gated Recurrent Units (GRU) in the generators and the predictors. Furthermore, they performed two synthetic experiments using BeerAdvocate dataset by deliberately inducing interlocking dynamics and then they performed experiments in real-world setting with BeerAdvocate and MovieReview. From the results they made two conclusions – 1) A2R showed consistent performance compared to other baselines on both the experiments, 2) A2R helps to promote trust and interpretable AI. In the future, the authors would like to improve A2R framework for generating casually corrected rationales to overcome the lack of inherent interpretability in the rationalization models. 

Existing methods in rationalization compute an overall selection of input features without any specificity and this does not provide a complete explanation to support a prediction. \citet{SG14} introduce ConRAT (Concept-based RATionalizer), a self-interpretable model which is inspired by human decision-making where key concepts are focused using the attention mechanism. The authors use the BeerReviews dataset to not only predict the review sentiment but also predict the rationales for key concepts in the review such as Mouthfeel, Aroma, Appearance, Taste, and Overall. ConRAT is divided into three subodels, a Concept Generator which finds the concepts in the review, a Concept Selector that determines the presence or absence of a concept, and a Predictor for final review predictions. ConRAT outperforms state-of-the-art methods while using only the overall sentiment label. However, \citet{ElsevierSub1} have further demonstrated that attention mechanism usage can contribute to a tradeoff between noisy rationales and a decrease in prediction accuracy.

\subsection{Text Classification}

\begin{table*}[h]
  \caption{Text Classification Papers}
  \label{tab:crtable}
  \resizebox{\textwidth}{!}{
  \begin{tabular}{cccccccc}
    \toprule
    Paper & Name & Year  & Explanation & Models & XAI Metric & Dataset & Code\\
    \midrule
   \cite{SG6} &   -  & 2007 & Extractive & SVM              & - & MovieReview                         &  - \\
\cite{SG7} &  -   & 2016 & Extractive & SVM, RA-CNN & - & Risk of Bias                        & -  \\
\cite{SG8} & GEF & 2019 & Extractive & CNN, LSTM        & - & PCMag Reviews, Skytrax User Reviews & \checkmark \\
\cite{SG9} & CDA & 2021 & Extractive & RL, RNN     & - & TripAdvisor Reviews, RateBeer       & \checkmark \\
    \bottomrule
  \end{tabular}}
\end{table*}

Text classification, also commonly known as text categorization, is the process of assigning labels or tags to textual data such as sentences, queries, paragraphs, and documents \cite{SG5}. Classifying text and extracting insights can lead to a richer understanding of the data but due to their unstructured nature, it is challenging and tedious. NLP techniques in text classification enable automatic annotation and labeling of data to make it easier to obtain those deeper insights of the data. We have found four papers in this field.

Traditionally, rationales provide well-defined kinds of data to nudge the model on why a prediction is the way it is given the data. Moreover, they require little additional effort for annotators and yield a better predictive model. When classifying documents, it is beneficial to obtain sentence-level supervision in addition to document-level supervision when training new classifications systems \cite{SG7}. Previous work relied on linear models such as SVMs (Support Vector Machines), therefore, \citet{SG7} propose a novel CNN model for text classification that exploit associated rationales of documents. Their work claims to be the “first to incorporate rationales into neural models for text classification”.  The authors propose a sentence-level CNN to estimate the probability that a sentence in a given document can be a rationale. They demonstrate that their technique outperforms baselines and CNN variants on five classification datasets. Their experimentation task uses Movie Reviews and the Risk of Bias (RoB) datasets. On the movie review dataset, their technique performs with a 90.43\% accuracy with the RA-CNN (Recurrent Attention Convolutional Neural Network) model and similar strong results are also indicated on the RoB datasets. 

It seems intuitive that more data or information can lead to better decision-making by the neural networks.  \citet{SG6} propose a new framework to improve performance for supervised machine learning by using richer “kinds” of data. Their approach is called the “annotator rationales” technique and it is to leverage a training dataset with annotated rationales. The rationales highlight the evidence supporting the prediction. \citet{SG6} test their approach on text categorization tasks, specifically, sentiment classification of movie reviews and they claim that these rationales enable the machine to learn why the prediction is the way it is. Rationales help the model learn the signal from the noise. ML algorithms face the “credit-assigment problem” which means that many features in the data (X) could have affected the predicted result (Y). Rationales provide a “shortcut” to simplifying this problem since they provide hints on which features of X were important. \citet{SG6} used a discriminative SVM for experimentation and the results indicate that this technique significantly improves results for the sentiment classification and they hypothesize that leveraging rationales might be more useful than providing more training examples. 

Recently, rationales have been a popular method in NLP to provide interpretability in the form of extracted subsets of texts. It is common to have spurious patterns and co-varying aspects in the dataset due to which rationale selectors do not capture the desired relationship between input text and target labels. Considering this problem this paper proposes CDA (Counterfactual Data Augmentation) framework to aid rational models trained with Maximum Mutual Information (MMI) criteria \cite{SG9}. CDA consists of transforms - for rational and classifications – because of their effectiveness over RNNs in NLP. The authors used TripAdvisor.com and RateBeer datasets for testing CDA with three baselines – MMI, FDA (Factual Data Augmentation), and ANT (simple substitution using antonyms). The results of the rational models were compared using precision and the accuracy of the classifier is reported based on the development set. From the results, authors concluded that the models trained using the CDA framework learn higher quality rationales and it doesn’t need human intervention. In the future, the authors would like to explore more on counterfactual predictors and on CDA framework that could connect with other rationalization strategies. Similarly, \citet{SG8} proposed a novel Generative Explanation Framework (GEF) for classification problems that can generate fine-grained explanations.  The motivation behind this explanation framework is to provide human-readable explanations without ignoring fine-grained information such as textual explanations for the label. For understanding the accuracy of explanations, the authors conducted experiments on two datasets - PCMag and Skytrax User Reviews – which were processed by the Stanford Tokenizer. Further, the authors used Encoder-Predictor architecture in which they used Conditional Variational Autoencoder (CVAR) as a base model for text explanations and Long Short-Term Memory (LSTM) for numerical explanations. The experimental results indicated that after combining base models with GEF the performance of the base model was enhanced along with improving the quality of explanations. Further, the authors also used human evaluation for evaluating the explainability of the generated text explanations. The authors noted that for 57.62\% of the tested items GEF provided better or equal explanations compared with the basic model.

\subsection{Neural Machine Translation}

\begin{table*}[h]
  \caption{Neural Machine Translation Papers}
  \label{tab:crtable}
  \resizebox{\textwidth}{!}{
  \begin{tabular}{cccccccc}
    \toprule
    Paper & Name & Year  & Explanation & Models & XAI Metric & Dataset & Code\\
    \midrule
   \cite{SG4} & SOCRAT & 2017 & Extractive & RNN         & Attention Score                                               & WMT14   & \checkmark \\
 \cite{SG3} &    -    & 2021 & Extractive & Transformer & LIME, Integrated Gradients & MLQE-PE & -  \\

    \bottomrule
  \end{tabular}}
\end{table*}

With the advent of deep learning, Neural Machine Translation (NMT) became the successor to traditional translation methods such as Rule-based or Phrase-Based Statistical Machine Translation (PBSMT) \cite{SG1}. NMT models leverage Deep Neural Networks’ architecture to train the model end-to-end to improve translation quality and only require a fraction of the storage memory needed by PBSMT models \cite{SG2}. The use of explanations to support NMT model’s prediction is relatively new, however, there has been some pioneering work to provide more explainability. We identified two relevant papers in this area.

Quality Estimation (QE) models perform well at analyzing the overall quality of translated sentences. However, determining translation errors is still a difficult task such as identifying which words are incorrect due to the limited amounts available training data. The authors explore the idea that since QE models depend on translation errors to predict the quality, using explanations or rationales extracted from these models can be used to better detect translation errors \cite{SG3}. They propose a novel semi-supervised technique for word-level QE and demonstrate the QE task as a new benchmark for evaluating feature attribution (the interpretability of model explanations to humans). Instead of natural language explanations, their technique employs various feature attribution methods such as LIME, Integrated Gradients, Information Bottleneck, causal, and Attention. It was shown that explanations are useful and help improve model performance and provide better explainability.

Deep learning models are black-boxes because they involve a large number of parameters and complex architectures which makes them uninterpretable. Considering this problem and to bring interpretability in deep learning models \citet{SG4} propose a model-agnostic method for providing explanations. The explanations provided by this method consist of sets of inputs and output tokens that are causally related in the black-box model. Further, these causal relations are inferred by performing perturbations on the inputs from the black-box models, generating a graph of tokens, and then solving a partitioning problem to select the most relevant components. To test the method authors used a symmetric encoder-decoders consisting of recurrent neural networks with an intermediate variational layer. This method was tested for three applications - simple mappings, machine translation, and a dialogue system. For simple mapping, the authors used the CMU (Carnegie Mellon University) Dictionary of word pronunciations and evaluated inferred dependencies by randomly selecting 100 key-value pairs. For Machine Translation the authors used three black-boxes – Azure’s Machine Translation system, Neural MY model, and human - for translating English to German. Finally, for the dialogue system, the authors used OpenSubtitle. From the results, the authors concluded that this model-agnostic method can produce reasonable, coherent, and often insightful expatiations. Additionally in future work, the authors noted that for Machine Translation and dialogue system applications potential improvements are needed for questioning seemingly correct predictions and explaining those that are not.

\subsection{Multiple Domains}

\begin{table*}[h]
  \caption{Multiple Domain Papers}
  \label{tab:crtable}
  \resizebox{\textwidth}{!}{
  \begin{tabular}{cccccccc}
    \toprule
    Paper & Name & Year  & Explanation & Models & XAI Metric & Dataset & Code\\
    \midrule
   \cite{SG17} & LIME    & 2016 & Extractive & SP-LIME, Parzen                               & - & Product Reviews & \checkmark \\
\cite{AK4} & ERASER  & 2020 & Extractive & LSTM, BERT-to-BERT                                     & Sufficiency                       & Movie Reviews, e-SNLI & \checkmark \\
\cite{SG19} & RGA     & 2020 & Extractive & Utility Function                                            &   -                                                   & Stockfish                                                               & -  \\
\cite{SG15} & EPITOME & 2020 & Extractive & RoBERTa, Attention                                          &     -                                                 & Redddit, TalkLife                                                       & \checkmark \\
\cite{SG18} & UNIREX  & 2021 & Extractive & BigBird-Base                                   & Comprehensiveness & SST, Movie Reviews, CoS-E                                               & -  \\
\cite{SG21} & ExPred  & 2021 & Extractive & MLP, GRU, BERT                                              &   -                                                   & FEVER, Movie Reviews                                           & \checkmark \\

    \bottomrule
  \end{tabular}}
\end{table*}

To demonstrate the effectiveness and generalizability of rationalization, many papers have attempted to demonstrate the use of rationales in multiple NLP tasks (\cite{SG15, SG16}) or in conjunction with other disciplines such as \cite{SG15}. In this section, we present six papers with work in more than one NLP task or if the work was in another discipline but leveraged rationalization. 

Currently in NLP many state-of-the-art tasks use deep neural networks and \citet{SG16} claim they are opaque in terms of their interpretability, or the way they make predictions. Lots of work has been conducted in this area, however, there is no standardization. The work has been with different datasets, NLP techniques and tasks which all have different aims and success metrics and this creates a challenge in this field of research in terms of tracking progress. To mitigate this, the authors propose a new benchmark called Evaluating Rationales And Simple English Reasoning (ERASER). There are multiple datasets (seven total) for various NLP tasks included in the benchmark. Datasets include human annotations of rationales which are the supporting evidence for a task’s prediction. This is an extractive rationalization technique. For example, ERASER includes the Movie Reviews dataset for sentiment classification and each review has a rationale or an annotated sentence that supports the prediction for that review. In addition, metrics are also provided as a baseline benchmark to evaluate the extract rationales quality. The authors believe that this benchmark will facilitate in creating better interpretable NLP architectures.

It is important to understand the reasons behind the predictions for assessing trust which is important for making decision or deploying a new model. Considering this problem, \citet{SG17} have proposed a novel model-agnostic approach LIME for providing explanations from any classifier about a local prediction and SP-LIME for providing global view of the model. For understanding the effectiveness of these methods simulated user experiment and evaluations with human subjects are performed. In the simulated user experiment two sentiment analysis datasets (books and DVDs) were used and different classifiers such as - decision trees, logistic regression (L2 regularization), nearest neighbors, and SVM with RBF (Radial Basis Function) kernel were also used. To explain individual predictions authors compared LIME with parzen, greedy and random procedures. Further, evaluation with human subjects is performed using Amazon Mechanical Turk to estimate the real-world performance by creating a new religion dataset. Based on the results the authors concluded that LIME explanations are faithful to the models and helps in assessing trust in individual predictions. Further, SP-selected LIME explanations are good indicators of generalization which is validated via evaluations with human subjects. In future authors would like to explore how to perform a pick step for images and would like to explore different applications in speech, video, and medical domains. Additionally, they also would like to explore theoretical properties and computational optimizations for providing accurate, real-time explanations which are needed in any human-in-the-loop systems.  

Rational extractions should be faithful, plausible, data-efficient, and fast with maintaining good performance but existing rational extractors are ignoring one or more of these aspects. Considering this challenge \citet{SG18} propose UNIREX (Unified Learning Framework for Rationale Extraction) an end-to-end rational extractor that accounts for all these mentioned aspects. For understating UNIREX’s effectiveness three text classification detests – SST (Stanford Sentiment Treebank), CoS-E (Commonsense Explanations), Movies – were used and compared with several baselines such as Vanilla, SGT (Sequence Graph Transform), SGT+P, FRESH etc. The authors used BigBird-base model in all their experiments which was pre-trained from the Hugging Face Transformers library. For training the authors used a learning rate of 2e-5 and batch size of 32 with maximum 10 epochs. For evaluating faithfulness, the authors used comprehensiveness (Comp) and sufficiency (Suff) for K = [1,5,10,20,50] which combined into a single metric Comp-Suff-Difference (CSD). For plausibility authors measured similarity to gold rationales using AUPRC (Area Under the Precision-Recall Curve), AP (Average Precision) and TF1 (Token F1). For data efficiency measurement how, performance varies with the percentage of train instances with gold rationale supervision is used. For speed authors compared methods via runtime complexity analysis with respect to the number of LM (Linear Model) forward/backward passes.  Finally for performance standard dataset-specific metrics accuracy (for SST and CoS-E) and macro-averaged F1 (for Movies) are used. The results indicated that UNIREX allows effective trade-off between performance, faithfulness, and plausibility to identify better rationale extractions. Further, authors also mention UNIREX trained rationale extractors can generalize to unseen datasets and tasks. 

With many ML systems demonstrating performance beyond that of human across many applications, the field of XAI is advancing techniques to improve transparency and interpretability. \citet{SG19} explores XAI in the context of a question previously unexplored in ML and XAI communities: “Given a computational system whose performance exceeds that of its human user, can explainable AI capabilities be leveraged to improve the performance of the human?”. The authors investigate this question through the game of Chess where computational game engines performance surpass the average player. They present an automated technique for generating rationales for utility-based computational methods called the Rationale-Generating Algorithm. They evaluate this with a user study against two baselines and their findings show that the machine generated rationales can lead to significant improvement in human task performance. They demonstrate that rationales can not only be used to explain the system’s actions but also instruct the user to improve their performance.

\citet{SG15} explores an application of rationalization in the mental health support field and understanding empathy. Empathy is important for mental health support and with the rise of text-based internet platforms, it becomes crucial to understanding empathy in only communication. The paper presents a computational approach to understanding empathy by developing a corpus of 10,000 pairs of posts and responses with supporting rationales as evidence. They use a multi-task RoBERTa-based bi-encoder model to identify empathy in conversations and extract rationales for predictions. Their results demonstrate that their approach can effectively identify empathic conversations.

To improve interpretability for NLP tasks, recent rationalization techniques include Explain-then-Predict models. In this technique, an extractive explanation from the input text is generated and then a prediction is generated. However, these models do not use the rationales appropriately and consider the task input as simply a signal to learn and extract rationales. \citet{SG21} propose a novel technique to prevent this problem with their approach  called ExPred where they leverage mult-task learning on the explanation phase and embed a prediction network on the extracted explanations to improve task performance. They experiment with three datasets (Movie Reviews, FEVER (Fact Extraction and VERification), MultiRC) and conclude that their model significantly outperforms existing methods.

\section{Extractive and Abstractive Methods}
\label{sec:extAbs}

\begin{center}
\includegraphics[scale=0.4]{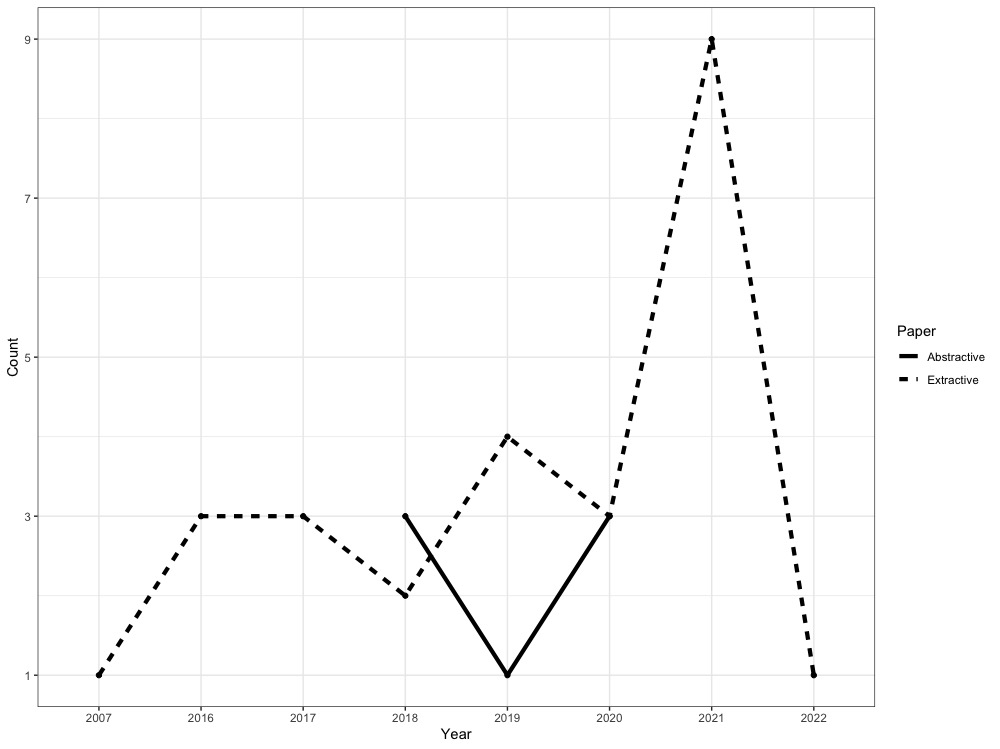}
\captionof{figure}{Papers Count by Type }
\label{fig:typescount}
\end{center}

This section compares extractive and abstractive rationalization techniques. It can be observed from Figure \ref{fig:typescount} that there is more interest and focus on extractive rationalization techniques compared to abstractive. There are multiple reasons for this, and the progress in the Automatic Text Summarization (ATS) domain can help explain.

\subsubsection{Extractive} In most extractive rationalization approaches, generating a rationale is similar to text summarization. These rationales contain the salient features of the input text, which users need to understand as the most influenced features of the model's prediction. 

Next, two steps are implemented while performing the task — i) irrelevant information is ignored, and ii) most crucial information is selected based on a scoring algorithm. This approach is a common foundation of summarization techniques. In extractive summarization, meaningful sentences are extracted to form a summary of the original text while still retaining the overall subject matter \cite{Disc1}. The critical difference with rationalization is that it is able to justify a neural network's prediction with evidence. In a way, extractive rationalization uses extractive summarization's fundamentals and takes it further. It frames the task as can we rationalize the output prediction where rationalize means to understand the prediction process and reason with supporting evidence. This introduces an interdependent relationship between the rationale and the prediction. This process is close to how humans rationalize with a sequence of reasons to justify a decision. This can be implemented in the NLP process to make models more explainable.

As interest in ATS systems grew in the past few decades, researchers have mainly focused on extractive summarization due to its simplicity, and reliability \cite{Disc1}. The abstractive summarization needed reliable natural language generation; thus, it was in its infancy from the 2000s to the early 2010s. Therefore, an increasing body of knowledge on extractive techniques is available, which researchers interested in rationalization could leverage and build on. This intuition behind extractive summarization paves the way for extractive rationalization.  The stark difference between extractive and abstractive in Figure \ref{fig:yearlypapers} is expected and reasonable, and the fields of summarization and rationalization follow similar paths. However, summarization approaches should purely be used for inspiration - following the identical methods for rationalization would be insufficient, and it does not provide reliable model explainability. \citet{SG18} notes that for appropriate explainability, the desiderata for the rationale is that - i) it must reflect the model's reasoning process (faithfulness), ii) be convincing to the nonexpert (plausibility), and iii) the rationale extraction should not hurt task performance. Thus, there is more work than simply extracting sentences as rationales. Moreover, extractive rationalization is insufficient because extracted sentences themselves are insufficient to provide full explainability. Humans do not fully understand without context and a coherent and logical explanation.

\subsubsection{Abstractive} The extensive research in extractive summarization reached its maturity, has peaked in terms of performance, and now the progress is stagnated \cite{Disc2}. Recent advances in deep learning and the advent of the Transformer architecture in 2017 have led to more reliable and influential language models \cite{Transformer}, \cite{BERTPaper}. In 2019,  \citet{Disc3} demonstrated a BERT-based abstractive summarization model that outperforms most non-Transformer-based models. Their model achieved state-of-the-art (SOTA) in automatic and human-based evaluations for summarization. Abstractive techniques allowed novel words and phrases to be generated instead of extracting spans from the input. Due to these advances, the research focuses gradually shifted from extractive to abstractive summarization. It is expected that rationalization will follow a similar trend. 

Abstractive rationalization is still relatively new, with limited research available. However, there have been promising and pioneering approaches such as \citet{AK9} and \citet{AK15}. Almost every paper discussed with an abstractive rationalization technique in Section \ref{sec:ratTech} leveraged some implementation of the Transformer architecture, such as BERT, GPT-2 \cite{GPT2Paper}, and T5, amongst others. BERT was the most frequently used language model. When BERT was released in 2018, it achieved SOTA results on many NLP tasks and surpassed human performance on tasks such as question answering and commonsense reasoning \cite{BERTPaper}. It made a giant leap in terms of performance compared to other language models of its time. This led to wide adoption and 
variations of BERT for the tasks where the Transformer-based model was required. Recently introduced models such as BART \cite{BARTPaper}, GPT-3 \cite{GPT3}, and T5 demonstrate promising results and surpass BERT in some tasks. This is due to language models growing exponentially, and they continue to improve and perform incredibly well at natural language generation \cite{DistilBERT}. For example, in some cases, text produced by GPT-3 is almost on par if not better, than human-written text. This enables more opportunities for research in abstractive rationalization, which is needed. By leveraging SOTA language models, explanations can become more comprehensive and convincing when illustrating a model's decision-making process. As mentioned in Section \ref{sec:background}, it is almost as if the models are “talking for themselves”. We believe that significant progress can be made in rationalization by focusing more on improving abstractive techniques.

\section{Discussions}
\label{sec:discuss}

In this section, we discuss insights from the literature reviewed, challenges, and potential future directions to propel progress on rationalization. Most importantly, we introduce a new XAI subfield called Rational AI.

\subsection{Introducing Rational AI}
\begin{center}
\includegraphics[scale=0.44]{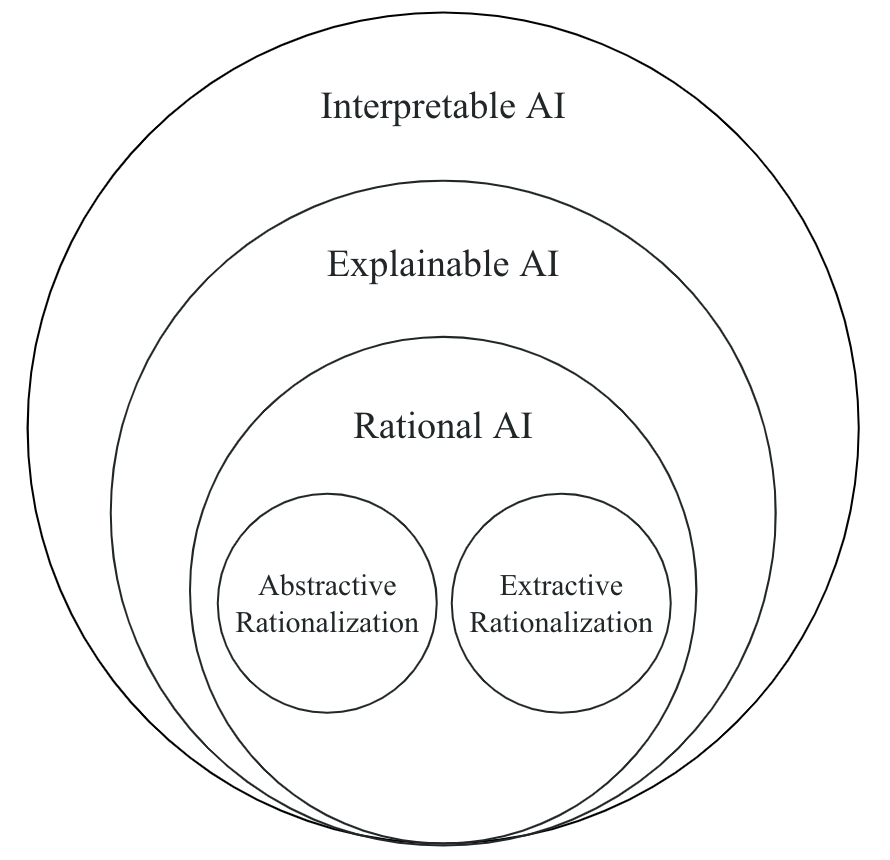}
\captionof{figure}{Rational AI}
\label{fig:rationality}
\end{center}

\subsubsection{Motivation} In Section \ref{section:intro}, we have seen the need for explainability and the available methods in NLP.  The numerical methods, such as SHAP values \cite{Disc4} or Attention scores, visualization methods, such as LIME \cite{SG17}, and saliency heatmaps, all require specialized domain knowledge to understand. 
At the same time, with increasing interactions with NLP-based systems, the nonexpert also deserves to know and understand how these black-box systems work because it has some degree of influence on their lives. This is formally called the \textit{right to an explanation}, a right to receive an explanation for an algorithm's output \cite{Disc6}. A classic example is a bank system with an NLP model that automatically denies a loan application. In this situation, providing the loan applicant with SHAP values or saliency heatmaps to justify the bank's algorithms is not very meaningful. Thus, explainability methods are truly explainable and helpful if the nonexpert can understand them \cite{Disc5}. We introduce Rational AI (RAI) as a potential solution.

\subsubsection{Rational AI} Rationalization techniques come the closest to this goal because they are built on natural language explanations (NLEs). NLEs are intuitive and human comprehensible because they are simply descriptive text. The textual information can be easily understood and translated into other languages if needed. Across all of the NLP tasks discussed in Section \ref{sec:ratTech}, we have seen the benefits of NLEs and the accessibility it provides to the nonexpert. We believe there is a critical need to focus on explainability techniques with NLEs. Considering these factors, we propose a new subfield in Explainable AI called Rational AI as shown in Figure \ref{fig:rationality}. We define Rational AI as follows. 

\vspace{0.5cm}

\noindent\textbf{Rational AI:} A field of methods that enable a black-box system to rationalize and produce a natural language explanation (rationale) to justify its output.

\vspace{0.5cm}

Rationality is the process of applying RAI to make models more explainable through an NLE. This is similar to the relationship between explainability and XAI. Further, rationality should not be confused or used interchangeably with the general AI term of a rational agent \cite{AITextbook}. These are distinct topics with similar names. In this survey, RAI and rationality are purely considered in the context of model explainability. We also have not seen any usage or previous definitions of RAI within this context. 

We compare rationality to the other fields shown in Figure \ref{fig:rationality}. Models with interpretability are interpretable, while those with explainability are interpretable and complete, as described in Section \ref{section:intro}. Models with rationality are interpretable and complete and can rationalize their behavior through an NLE. 


The explainability methods described earlier in this subsection explain, but they do not justify in a way that is accessible and comprehensible to the nonexpert. In recent years, language models have become powerful and incredibly good at language generation tasks, but we have yet to see their full potential. As they continue to grow exponentially, we predict this is the beginning of explainability techniques using NLEs. The intuition behind RAI is that rationalization is one such technique, and many are yet to be developed. This calls for a more organized field to improve research focus and the need for RAI to exist.

\subsubsection{Generalizing RAI} Although RAI arises from the need for better explainability for NLP tasks, it is potentially applicable in general AI and other fields in AI. Other fields, such as Computer Vision, Speech, and Robotics, could leverage rationalization methods to improve their model explainability. For example, rationalization in Computer Vision can help explain through an NLE which visual features contributed the most to an image classifier prediction in place of complex explainable techniques \cite{IntegratedGradients, ElsevierSub7}. Many promising opportunities exist for researchers to apply rationalization in other disciplines.

\subsection{Challenges}
We have seen that rationalization is a relatively new technique, and with it, various challenges exist. In this subsection, we share challenges and potential solutions to improve the current state.

\subsubsection{Statistical Evaluations}  No standard statistical evaluations exist currently for rationalization. There is a wide variety of metrics that are in use, such as Mean Squared Error \cite{SG22}, Accuracy \cite{AK9, SG6, SG11}, F1 Score \cite{AK26, AK23}, ANOVA (Analysis of variance) \cite{SG19}, and Precision \cite{SG9}. We have observed that the most preferred statistical metric is accuracy. It is reasonable for evaluation metrics to be task-dependent and focused on the prediction. However, those alone are insufficient because the accuracy of the NLE also needs to be considered. For example, if the task prediction had high accuracy, but the NLE was unclear and incomprehensible, then it is not helpful. Metrics such as the BLEU (BiLingual Evaluation Understudy) score by \citet{BLEU} and the ROUGE-N (Recall-Oriented Understudy for Gisting Evaluation) score by \citet{ROUGE} exist for evaluating open-ended machine-generated texts. However, we have seen limited use in the literature review, such as \cite{AK13}. The scores work by comparing the generated text with a set of ground-truth reference texts, and often these are human-written references. These scores are helpful, especially for abstractive rationalization, where explanations can be open-ended; however, they come with limitations since the evaluation is effectively token-level matching. Since an NLE is the typical outcome of systems with rationalization, adopting a standard evaluation metric can help improve research progress. Consistent evaluations also make it easier to compare different experiments and approaches.

\subsubsection{Data}The availability and the need for more diversity of appropriate datasets is also a problem hindering progress.

Availability: Data collection is an expensive and time-consuming task. It is possible to repurpose existing datasets, but modifying them requires manual human labor. Thus, researchers often build their datasets for a specific task they are working on. \citet{AK13} developed the e-SNLI dataset by modifying the SNLI dataset from \citet{SNLI}. \citet{AK13} achieved promising results on their task, demonstrating how their dataset can enable a wide range of new research directions by altering and repurposing existing datasets.

Diversity: Without enough datasets, new research in rationalization will be limited. Researchers will be constrained to the existing datasets to make new progress. This trend is evident in the literature reviewed in MRC and Sentiment Analysis compared to NMT. In MRC, the datasets are very diverse. In sentiment analysis, most papers rely on either the BeerAdvocate \cite{BeerAdvocate} or MovieReviews \cite{MovieReviews} datasets to perform their experiments. In both domains, we discovered five publications each. For a domain such as NMT, progress seems limited, and we found only two publications. The lack of appropriate rationalization datasets for NMT tasks is a possible reason for this. 

As we observed in our literature review, there is a direct relationship between dataset availability and the progress made. More work in creating new datasets for rationalization can help improve diversity and the progress of certain domains lagging behind, such as NMT. New datasets across all domains, in general, will increase the interest and work in rationalization because researchers will have more flexibility in designing new techniques and experimenting with a wide variety of data. \citet{DataShop} has organized the largest repository of learning science datasets called DataShop, and it led to improvements in research progress. Similarly, an organized central repository for rationalization supporting datasets can be beneficial.
Without a centralized model evaluation and development system, reproducibility and accessibility will remain low.

\subsection{Human-Centered Evaluations and Assurance}
NLP has direct applications in many disciplines. For example, MRC and commonsense reasoning are helpful in the education discipline. Our literature review indicates using Q\&A tools and commonsense injection to generate explanations for educational needs \cite{AK2}, \cite{AD2}. Further, NLP has also been used to enhance human task performance, as we saw in \cite{SG19}, and to provide support for mental health \cite{SG15}. Additionally, fact-checking is another application, and it is crucial in social media, fake news detection, and law \cite{AK23}. It has become common to interact with these systems, and they may have a significant influence on all aspects of our society. Due to this, the European Union recently passed a regulation that requires algorithms to provide explanations that can significantly affect users based on their user-level predictions \cite{Intro3}. 

\subsubsection{Human-Centered Evaluations (HCE)}The explanations provided by the NLP systems must provide enough information to the user to help them understand its decision-making process \cite{AD7}. Considering these aspects, the human-machine partnership is essential for evaluating and generating accurate explanations. This calls for better methods to evaluate the explanations generated. The field of HCE addresses this problem, and \citet{AD8} defines it as a "field of research that considers humans and machines as equally important actors in the design, training, and evaluation of co-adaptive machine learning scenarios."

In this literature survey, we found 15 out of 33 papers in which HCE is performed. \citet{AD8} mentions that there has been an increasing trend of HCE since 2017 compared to the previous years. While conducting this literature survey, this trend was not observed in the rationalization domain. Overall, we found that HEC is incorporated in most of the papers on Machine Reading Comprehension (2 out 5), Commonsense Reasoning (3 out 4), and multiple domains (4 out 6). From our observations, researchers give more attention to performance while evaluating AI algorithms and ignore human factors such as usability, user intentions, and user experience. Thus, along with the accuracy of AI algorithms, it is also essential to focus on the interpretability and reliability of the explanations generated by AI algorithms. The articles in which HCE is used are primarily performed via crowdsourcing using Amazon Mechanical Turk, and the focus is on user-based evaluations or annotations. This pattern necessitates conducting expert evaluations to understand users' needs better because it can help improve trust in AI algorithms. 

\begin{center}
\includegraphics[scale=0.27]{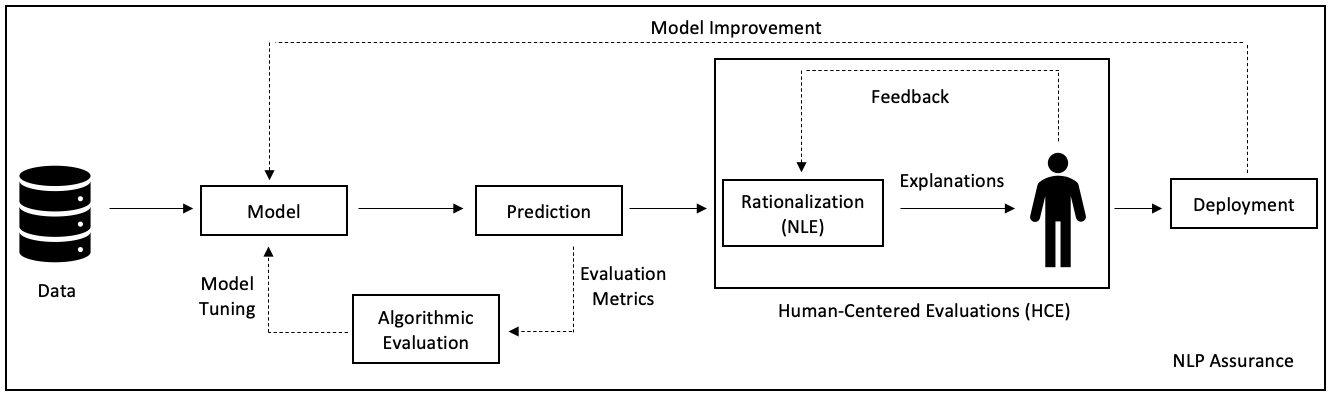}
\captionof{figure}{Integration of HCE to enable NLP Assurance.}
\label{fig:nlplifecycle}
\end{center}

HCE is a subset of the Human-Computer Interaction (HCI) field, which is integrated with the AI paradigm after the algorithmic performance evaluations as shown in Figure \ref{fig:nlplifecycle}. This integration can be regarded as human-centered AI, and \citet{AD9} claims this as an AI/ML perspective that intelligent systems are part of a more extensive system that also includes human stakeholders. The literature \cite{AD10} on usability, and user experience testing demonstrated three widely used methods to perform HCE - Think Aloud (TA), Heuristic Evaluation (HE), and Cognitive Walkthrough (CW). The TA method is a standard method and can be more effective considering the evaluations of explanations in the NLP domain. In the TA method, evaluators are asked to "think aloud" about their experience while an experimenter observes them and listens to their thoughts \cite{AD11}. This way, an HCE method can be used in the final step to understand usability, user intentions, and user experience. This may lead to a better understanding of the interpretability and reliability of the explanations generated by rationalization. Therefore, in addition to statistical evaluation techniques, we strongly encourage researchers to integrate HCE as part of their evaluations.

\subsubsection{Assurance} Further, these checks may be crucial to enable trustworthy and transparent NLP systems to achieve NLP Assurance. It is critical to perform rigorous testing and validation of NLP systems at all stages before their deployment. For example, it should be ensured that the data is unbiased, models are interpretable, and the process of arriving at the outcome is explainable to a nonexpert. In the last step of this process, it would be beneficial to use RAI techniques. Integrating rationalization with human-centered evaluations and elements of NLP Assurance can invoke human-AI trust and safety with the systems. This process may also transform black-box systems into white-box systems and make NLP models more comprehensible and accessible for nonexpert users.

\section*{Acknowledgments}
We would like to acknowledge the members of the A3 lab (https://ai.bse.vt.edu/) at Virginia Tech for their support and input to this work.

\section*{Declarations}

\subsection*{Funding}
This work was supported by the Commonwealth Cyber Initiative (CCI).

\subsection*{Competing interests}
The authors declare that they have no competing interests.

\subsection*{Ethics approval and consent to participate}
Not applicable.

\subsection*{Consent for publication}
Not applicable.



\bibliographystyle{ACM-Reference-Format}
\bibliography{references}

\end{document}